# Brain segmentation based on multi-atlas guided 3D fully convolutional network ensembles

Jiong Wu[a,b], Xiaoying Tang[b,*]

a. School of Electronics and Information Technology, Sun Yat-sen University, Guangzhou, Guangdong, China

b. Department of Electrical and Electronic Engineering, Southern University of Science and Technology, Shenzhen, Guangdong, China

## Abstract

In this study, we proposed and validated a multi-atlas guided 3D fully convolutional network (FCN) ensemble model (M-FCN) for segmenting brain regions of interest (ROIs) from structural magnetic resonance images (MRIs). One major limitation of existing state-of-the-art 3D FCN segmentation models is that they often apply image patches of fixed size throughout training and testing, which may miss some complex tissue appearance patterns of different brain ROIs. To address this limitation, we trained a 3D FCN model for each ROI using patches of adaptive size and embedded outputs of the convolutional layers in the deconvolutional layers to further capture the local and global context patterns. In addition, with an introduction of multi-atlas based guidance in M-FCN, our segmentation was generated by combining the information of images and labels, which is highly robust. To reduce over-fitting of the FCN model on the training data, we adopted an ensemble strategy in the learning procedure. Evaluation was performed on two brain MRI datasets, aiming respectively at segmenting 14 subcortical and ventricular structures and 54 brain ROIs. The segmentation results of the proposed method were compared with those of a state-of-the-art multi-atlas based segmentation method and an existing 3D FCN segmentation model. Our results suggested that the proposed method had a superior segmentation performance.

## 1 Introduction

Brain segmentation is one of the most important steps in studying the complex relationships between specific anatomical structures and various brain disorders such as Alzheimer's disease (Iglesias et al., 2016; Tang et al., 2016), Parkinson's

Abbreviations: ROI (region of interest), CNN (convolutional neural network), S-FCN (single fully convolutional network), M-FCN (multiple fully convolutional network), JLF (joint label fusion) disease (Geevarghese et al., 2015) and Multiple sclerosis (Garcia-Lorenzo et al., 2013). Although manual delineation is usually believed to be the most accurate, it suffers from being time-consuming and of large intra- or inter- operator variability (Cavanaugh et al., 2002). With the advent of large-scale neuroimaging studies, developing accurate, robust and fully-automatic brain segmentation techniques is extremely urgent in quantitative magnetic resonance imaging (MRI) analysis.

In the past decade, many semi- and fully- automatic approaches have been proposed to segment anatomical regions of interest (ROIs) from brain MR images. One of the main streams is atlas-based methods, which relies on nonlinear image registrations (Babalola et al., 2009; Collins et al., 1995) and pre-defined atlases. Due to the ability of integrating various expert priors, atlas-based segmentation methods have been widely used. Single atlas-based methods may be incapable of capturing inter-subject variability (Tang et al., 2013). In such context, multi-atlas based methods relying on majority voting have been proposed (Aljabar et al., 2009; Heckemann et al., 2006; Rohlfing et al., 2004; Warfield et al., 2004). Although these approaches have incorporated information from multiple atlases that are similar to the to-be-segmented image, they are sensitive to registration errors. To address this issue, non-local patch-based label fusion methods have been proposed (Coupe et al., 2011; Rousseau et al., 2011). Existing patch-based methods usually rely on affine but not nonlinear registrations. They first affinely align multiple atlas images to the to-be-segmented target image, and then assign weights to the transformed atlas labels according to patch similarity. These patch-based methods were then improved by combining with other techniques, such as a multi-resolution framework (Eskildsen et al., 2012), active appearance models (Hu et al., 2014) and level sets (Wang et al., 2014). Wang and Yushkevich (2013) further proposed a joint probability model to reduce the inconsistency among nonlinear registrations in terms of the transformed atlases.

Although the aforementioned patch-based methods can effectively reduce the dependency on precise registrations, patches of fixed size lack discriminative power to characterize complex appearance patterns of different ROIs. To alleviate this limitation, Wu et al. (2015) proposed a new label fusion strategy by incorporating multi-scale feature representation and label-specific patch partition, which can capture certain complex appearance patterns of different ROIs and exclude dissimilar patches during the weighting process. A common drawback of patch-based methods is that the process of searching similar patches is highly computationally expensive.

In addition to multi-atlas based and patch-based segmentation methods, learning-based methods using discriminative features for label prediction have also been explored, usually in a patch-based manner. For example, Tu and Bai (2010) employed a probabilistic boosting tree, with context features extracted from the atlases, to train a classifier for brain labeling. Tong et al. (2013) proposed a segmentation method using sparse coding to learn the dictionary. Zikic et al. (2014) proposed a method for multi-atlas label propagation using random forest. Hao et al. (2014) used L1-regularized support vector machine to segment the hippocampus. Moreover, employing the 3D Haar-like features and an auto-context model makes the learning-based segmentation methods more robust (Wu et al., 2018). These approaches however largely depend on features extracted and usually involve heavy algorithm design.

With the development of artificial neural networks, deep learning has been suggested to be a powerful alternative for supervised learning. Different from the traditional learning methods that use specifically-designed features, deep learning techniques can automatically identify hierarchical features corresponding to different levels of abstraction of the original data. Among different deep learning architectures, convolutional neural networks (CNNs) (Krizhevsky et al., 2012a; Lecun et al., 1998) have shown outstanding performance in recognition tasks, especially in image classification.

Because of its superior performance in recognition tasks, CNNs have recently been introduced into medical image segmentation. Ciresan et al. (2012) trained a CNN in a sliding-window setup to predict the class label of each pixel with the patch surrounding that pixel being the input. That CNN, however, has two drawbacks including slow segmentation speed and high sensitivity to patch size. Inspired by this work, some other CNNs have been proposed for glioblastoma tumor segmentation. For instance, Pereira et al. (2015) presented a deeper CNN architecture using small kernels with intensity normalization for segmenting brain tumor, being the second best performing approach in the BRATS 2015 challenge (Menze et al., 2015). Havaei et al. (2017) used a two-path CNN architecture, with one path being used for capturing local details and another for capturing global context. Some recent studies have also investigated CNNs for segmenting brain ROIs. For instance, Zhang et al. (2015) proposed a deep CNN for segmenting brain tissues (white matter, gray matter, cerebrospinal fluid) using multi-modality MR images. Moeskops et al. ( 2016) presented a CNN model trained by a large amount of multiscale patches for segmenting brain structures using five different datasets. These methods were based on 2D CNNs, which had the advantages of low memory requirement, fast computation, and easy transfer from existing architectures obtained from computer vision tasks. However, 2D CNNs lack the ability to exploit dense inference (Lecun et al., 1998; Sermanet et al., 2013). To address this limitation, several 3D CNNs have been proposed. Kamnitsas et al. (2017) proposed a 3D CNN architecture for brain tumor segmentation. Similar to the work in Havaei et al. (2017), this approach used a two-path architecture, with the first one receiving a subregion of the original to-be-segmented image and the second one receiving a much larger region from the down-sampled to-be-segmented image. Fang et al. (2017) proposed a multi-atlas guided 3D fully convolutional network (FCN) for brain image labeling. To reduce computation and memory load, Dolz et al. (2018) proposed a deep 3D FCN using small kernels for subcortical segmentation without any need for image registration.

All of the recently proposed architectures that are related to this work used patches of fixed size, which obstructed capturing global context features of different ROIs. They may generate isolated label prediction and false segmentation. Furthermore, these existing methods (Dolz et al., 2018; Moeskops et al., 2016; Shakeri et al., 2016) did not employ image registration, and label prediction relied solely on the local appearance of image patches, which may impair the segmentation accuracy. Although the method proposed in Zhang et al. (2017) has introduced label information into the FCN model, its learning ability was limited by the simple structure of its FCN. To address the limitations of these recently-developed 3D FCN architectures for brain segmentation and to combine with the advantages of multi-atlas based methods, in this paper, we proposed and validated a multi-atlas guided 3D FCN ensemble model (M-FCN) for segmenting brain ROIs. The major contributions of this work are three-fold: 1) We proposed an adaptive size determination method for the extraction of different ROI patches; 2) The proposed model is efficient in propagating expert priors by inputting both the image and the label information of several most similar ROI patches into 3D FCNs for training and testing; 3) The proposed 3D FCN is an end-to-end network including only convolutional and deconvolutional layers which can be efficiently trained in less than 10 epochs.

In this work, we conducted comprehensive validation experiments on two brain MRI datasets, with the first one focusing on 14 subcortical and lateral ventricle structures and the second one focusing on whole brain segmentation involving a total of 54 brain ROIs except brainstem and cerebellum. The first dataset was used to compare the proposed M-FCN with the multi-atlas guided single FCN model (S-FCN), a representative patch-based multi-atlas method (Wang and Yushkevich, 2013), and a representative 3D FCN method (Dolz et al., 2018). The second dataset was used to compare M-FCN with S-FCN and the patch-based multi-atlas method (Wang and Yushkevich, 2013) given that the selected 3D FCN method (Dolz et al., 2018) was specifically designed and tested for subcortical structures.

# 2 Methods

In section 2.1, we detail the definition of the segmentation problem and the overall framework of the proposed segmentation model. Section 2.2 describes the method for determining the patches' centers and sizes of different ROIs. The proposed multi-atlas guided 3D FCN architecture, comprised by encoding and decoding layers, will be presented in section 2.3. In that section, we also show how this architecture can be improved by combining multi-scale features across different layers. Thereafter, section 2.4 presents the model of ensemble 3D FCNs and describes how the single 3D FCN architecture can be improved by ensemble techniques. The segmentation pipeline will also be described in section 2.4. Finally, section 2.5 focuses on the study design and experimental setup, providing information on the datasets used in this study, implementation details of the testing network architectures, and the evaluation criterion of each segmentation method's performance.

## 2.1 Problem definition

Let $I_T$ be a to-be-segmented image $I_T = \{I_T(x) | x \in \Omega\}$, where $x$ denotes each image voxel. The goal of a multi-atlas based segmentation method is to estimate a label map $L_T$ which assigns a label $l_x \in \{1, \ldots, N\}$ ($N$ denotes the total number of brain structures of interest) to each voxel of the to-be-segmented image, given $M$ atlases $A_1, \ldots, A_M$ with $A_M = (I_M, L_M)$ where $I_M$ and $L_M$ are the grayscale image and the corresponding label image of the atlas $A_M$. In a general CNN, features will be automatically extracted in the convolutional layers. The last layer will usually use the softmax function to get the probabilities that the input belongs to each of the output classes. The softmax function at a specific voxel $x$ can be expressed as

$$S_{x_j} = \frac{\exp(a_j)}{\sum_{k=1}^{N} \exp(a_k)} \quad \forall j \in 1 \ldots N, \tag{1}$$

where $N$ denotes the total number of output classes, $a$ is a vector of size $N$ generated from preceding layers. The segmentation problem is then solved via

$$\hat{l}_x = \underset{j}{\operatorname{argmax}}\, S_{x_j}. \tag{2}$$

In our segmentation framework, we process each ROI independently. Hence, the multi-class problem is decomposed into multiple binary-class problems, yielding

$$\hat{l}_{x_j} = \begin{cases} 0, S_{x_j} \leq 0.5 \\ 1, S_{x_j} > 0.5 \end{cases}. \tag{3}$$

Finally the multi-class problem can be solved via a fusion estimation

$$\hat{l}_x = \underset{j}{\operatorname{argmax}}\, \hat{l}_{x_j} \cdot S_{x_j} \ . \tag{4}$$

**Figure 1** illustrates the proposed segmentation framework. At the beginning of the training stage, histogram matching is performed between a randomly selected image and all remaining images. After that, $m$ $(0 < m < M)$ images and labels from the $M$ atlases are selected and considered as the template atlases, and the remaining $n = M - m$ atlases are considered as the target atlases. Each template image is aligned to each target image using an affine registration followed by a fast version of the large deformation diffeomorphic metric mapping (LDDMM) (Wu and Tang, 2018). Then the size of each ROI patch is determined adaptively and the centers of the training patches are sampled from the target images. Next, for each ROI, we extract multiple 3D training atlas patches (including both the image patches and the corresponding label patches) from the target images and labels, and for each training patch, we identify $K$ most similar atlas patches from the aligned template atlases. Then, each of the extracted training atlas patches and the corresponding $K$ similar atlas patches are fed into the ROI-specific M-FCN model as a training sample.

At the testing stage, the same preprocessing steps including histogram matching and affine+LDDMM registration between the testing image and each of the template images are conducted. For each ROI, the center of the testing image patch is then calculated from the deformed template labels. With a determination of the patch size at the training stage and the center at the testing stage, the testing image patch as well as $K$ most similar atlas patches from the aligned template atlases are extracted. The extracted testing image patch and the $K$ most similar atlas patches are finally loaded into the trained ROI-specific M-FCN model to obtain the predicted label patch and probability map. After conducting the same procedure for all ROIs, all label patches and the corresponding probability maps are obtained. Then, label fusion is performed to yield the final segmentation result.

Unlike traditional CNNs, FCNs are only composed of convolutional layers allowing them to be applied to images of arbitrary size (Long et al., 2015). As such, in our work, we use a similar FCN architecture for learning all ROIs, although different ROIs may have different patch sizes.

## 2.2 Patch Extraction

In the proposed method, we use patches of adaptive sizes to train the FCN networks. **Figure 2** illustrates the method for determining the patch size and the center of a specific ROI in a testing image. For each target image, the transformed template labels are obtained by applying the transformations from the affine+LDDMM registrations to the original template labels. For the $i$th ROI of the $j$th target $\mathcal{R}_{ij}$, a coordination area $\mathcal{A}_{ij}$ is defined as,

$$\mathcal{A}_{ij} = \mathcal{L}_1 \cup \mathcal{L}_2 \cup \cdots \cup \mathcal{L}_m, \tag{5}$$

where $\mathcal{L}_k$ $(k = 1, \ldots, m)$ denotes the coordinates of the specific ROI in the $k$-th transformed template label image. The region $\mathcal{A}_{ij}$ is then dilated to generate an expanded region $\mathcal{A}'_{ij}$. Then, a cuboid $\mathcal{C}_{ij}$ which exactly contains $\mathcal{A}'_{ij}$ is obtained. Let $r_{x_{ij}}$, $r_{y_{ij}}$ and $r_{z_{ij}}$ respectively denote the length, width and height of $\mathcal{C}_{ij}$, then the size of patch $\mathcal{C}_i$ for the $i$th ROI, namely $(r_{x_i}, r_{y_i}, r_{z_i})$, is computed as,

$$(r_{x_i}, r_{y_i}, r_{z_i}) = \left(max(r_{x_{ij}}), max(r_{y_{ij}}), max(r_{z_{ij}})\right). \tag{6}$$

Because of the high accuracy of affine+LDDMM registration (Wu and Tang, 2018), a reasonable radius of dilation will always make $\mathcal{R}_{ij} \subseteq \mathcal{A}'_{ij}$, and thus $\mathcal{R}_{ij} \subseteq \mathcal{C}_i$. To determine the patch center of a specific ROI in a testing image, we conduct an union operation of the $m$ corresponding transformed template labels and a dilation operation to generate a cuboid, and then the center $O(x, y, z)$ of this cuboid is considered as the ROI patch's center. Finally, given the patch center and the patch size, the ROI patch of the testing image can be extracted.

To train an M-FCN model, we sample each training atlas patch and the corresponding $K$ most similar atlas patches respectively from the original target atlas and the corresponding transformed template atlases. For each training image patch $P_{T(I,j)}$ centered at voxel $j$ and extracted from the target image $I$, similar to the method in Fang et al. (2017), we search the most similar image patch from each transformed template atlas in a 3D cubic searching neighborhood, resulting in a total of $m$ atlas patches. The searching neighborhood is a bounding box centered at voxel $j$ and denoted as $c(j)$. The similarity between two patches is quantified according to the sum of intensity differences

$$\hat{P} = \left\{ P_{A(m,o)} \mid \min_{o \in c(j)} \left\| P_{T(I,j)} - P_{A(m,o)} \right\|_2^2 \right\}, \quad (7)$$

where $P_{A(m,o)}$ denotes the patch in the searching neighborhood. Then we rank all of the selected $m$ atlas patches using the patch similarity measure and identify the top $K$ atlas patches. The patch size for the $i$-th ROI is $\boldsymbol{R} = (r_{x_i}, r_{y_i}, r_{z_i})$ throughout the entire patch extraction pipeline. **Figure 3** summaries all steps involved in extracting the training patches of each ROI.

## 2.3 The proposed multi-atlas guided 3D FCN architecture

A "fully convolutional" approach is used in this work. **Figure 4** depicts the overall framework of the proposed multi-atlas guided 3D FCN model which consists of a multi-encoding part and an encoding-decoding part.

Denote the training data as $\boldsymbol{S} = \{(\boldsymbol{X}_n, \boldsymbol{Y}_n),\ n = 1,2, \dots, \mathcal{N}\}$, where $\boldsymbol{X}_n = \left\{ \left( x_n^j, p_{n1}^j, \dots, p_{nk}^j, l_{n1}^j, \dots, l_{nk}^j \right),\ j = 1,2, \dots, \mathcal{M} \right\}$ denotes the $n$-th training sample including one training image patch $\boldsymbol{x}_n$, $K$ most similar transformed template image patches $\boldsymbol{p}_n$ and the corresponding $K$ transformed label patches $\boldsymbol{l}_n$, $\boldsymbol{Y}_n = \left\{ y_n^j, j = 1, \dots, \mathcal{M} \right\}$ denotes the segmentation ground truth (binary mask) of the $n$-th training image patch, $\mathcal{N}$ denotes the total number of training samples and $\mathcal{M}$ denotes the total number of voxels in the $n$-th training patch. For simplicity, we denote all parameters in the proposed FCN as $\boldsymbol{\theta}$, $\boldsymbol{\theta} = \{\boldsymbol{W}, \boldsymbol{b}\}$, where $\boldsymbol{W}$ denotes weights and $\boldsymbol{b}$ denotes bias of the FCN. With the predicted segmentation maps being $\mathfrak{L}(\boldsymbol{\theta}, \boldsymbol{X}_n)$, the goal of the training process is to minimize an objective function to get the optimal parameters $\widehat{\boldsymbol{\theta}}$,

$$\widehat{\boldsymbol{\theta}} = \underset{\boldsymbol{\theta}}{\operatorname{argmin}} \frac{1}{\mathcal{N}} \sum_{n=1}^{\mathcal{N}} L(\mathfrak{L}(\boldsymbol{\theta}, \boldsymbol{X}_n), \boldsymbol{Y}_n), \tag{8}$$

where $L(\cdot,\cdot)$ denotes the cost function.

In the multi-encoding part, the training image patch and the $K$ most similar atlas patches are separately learned. Specifically, as depicted in **Figure 4**, each transformed template image patch and the corresponding label patch are concatenated in a multi-channel manner. Then, after 2 convolutional layers and 1 max-pooling layer, all the $K+1$ patches are concatenated and fed to the encoding-decoding part. This multi-encoding (multi-atlas guided) strategy has two important advantages over other single patch learning approaches. On the one hand, it introduces the ROI's label information into the network. On the other hand, the concatenation of $K+1$ patches ensures a more relevant context around the specific ROI which improves the robustness of the segmentation.

In the encoding-decoding part, a 3D FCN including 7 convolutional layers and 10 deconvolutional layers are employed. Each convolutional layer contains several 3D convolutional filters (or kernels). With the number of convolutional kernels in layer $l$ being $m_l$, the $n$-th input to layer $l$ being $x_{l-1}^{n}$, the $k$-th output feature map of layer $l$ is computed as:

$$y_l^k = f\left(\sum_{n=1}^{m_l-1} W_i^{k,n} \otimes x_{l-1}^{n} + b_l^k\right), \tag{9}$$

where $W_i^{k,n}$ denotes the filter convolved with each of the previous layers, $\otimes$ denotes the convolution operator, $b_l^k$ denotes the bias, and $f$ denotes the nonlinear activation function.

In addition to these two parts, this architecture also combines the feature maps of various intermediate layers. As shown in **Figure 4**, three feature maps including the output of the 3rd layer in multi-encoding, the outputs of the 3rd and the 6th layer in encoding-decoding are respectively concatenated to the outputs of the15th, the 12th and the 9th layer of the encoding-decoding part. This strategy has two important advantages over other approaches (Kamnitsas et al., 2017, Dolz et al., 2018, Fang et al., 2017): 1) The concatenation operation improves the FCN's ability in spatial information learning. 2) Since the sizes of the feature maps of these two connected layers are almost the same, the spatial context can be efficiently combined without information loss.

Throughout the network, zero-padding is used so that the size of the feature maps is only changed by strided convolutional or deconvolutional operations. Each activation layer in the multi-encoding part and encoding of the encoding-decoding part is preceded by a layer of batch normalization. To reduce overfitting, dropout strategy is used in all layers.

## 2.4 Ensemble FCNs

Ensemble learning is an extensively studied technique in machine learning and pattern recognition, especially in dealing with small sample size, high-dimensionality and complex data structure problems. Ensemble learning is a machine learning paradigm where multiple learners are trained to improve the performance of the overall system. Representative ensemble techniques include multi-model combination learning, multi-strategy learning, classifiers fusion and so on (Tan and Gilbert, 2003; Yang et al., 2010). In this paper, we use multi-model combination learning and multi-strategy learning to improve the segmentation results. The motivations are two-fold: 1) To reduce over-fitting, the dropout technique used in S-FCN model can generate different FCN architectures enabling the corresponding models learn different attributes of the training data, and thus an ensemble of them may boost the segmentation results; 2) Randomness strategies such as a random initialization of parameters and a random shuffling of training samples may cause the network to get stuck at different local minima. As such, using randomness strategies for multiple times in multiple models may reduce this impact.

As illustrated in **Figure 5**, $n$ S-FCN models with different numbers of neurons in each hidden layer are trained with randomly shuffled data. At the testing stage, given a testing sample, $n$ probability maps will be generated using the $n$ S-FCNs. After averaging those $n$ probability maps, the binary segmentation map can be obtained using formula (3). The average probability segmentation maps of $N$ ROIs, as well as the binary maps are plugged into formula (4) to yield the final segmentation map for the testing image.

## 2.5 Study design and experiment setup

### 2.5.1 Datasets

Two datasets were used for testing the proposed segmentation method. The first dataset came from the PREDICT- HD study (https://www.predict-hd.net/). This dataset was employed to quantify the performance of four different FCN architectures, and compare the performance of the proposed method with another two state-of-the-art segmentation methods. There are a total of 16 subjects including 3 males and 13 females (mean age = 42.1 (10.1 years), resulting in a total of 16 high resolution, T1-weighted 3D-volume MPRAGE images. For each of the 16 images, a total of 14 ROIs have been manually delineated, including the left and right caudate, putamen, globus pallidus, thalamus, hippocampus, amygdala, and lateral ventricle. All volumes have a size of $190 \times 230 \times 180$ voxels. To ensure an unbiased evaluation of the segmentation performance, a 4-fold cross-validation strategy was employed, wherein each fold consisted of 10 training images (4 images were randomly selected as the template, and the other 6 as the target), 2 validation images and 4 testing images.

The second dataset is publicity available, known as LPBA40, containing 40 brain images (20 males and 20 females, mean age = 29.20 (6.30 years). These images were acquired at the Laboratory of Neuro Imaging at the University of Southern California and freely available (http://www.loni.usc.edu/atlases/). This dataset originally contains expert-labeled segmentations of 56 ROIs. We excluded brainstem and cerebellum from our analyses, resulting in a total of 54 ROIs for this dataset. We also used a 4-fold cross-validation strategy. In each fold, 28 images were used for training involving 10 randomly selected template images and 18 target images, 2 images for validation and 10 images for testing.

### 2.5.2 Implementation details

Given that a specific ROI in different images generally varies in terms of both size and location, to ensure an ROI is completely contained in the extracted image patch, image registration operations are performed before the patch extraction. In this work, we employed an affine followed by LDDMM to align the template images to the target and the testing images. To be specific, a 12-parameter affine alignment via AIR (Woods et al., 1998a; Woods et al., 1998b) followed by a three-cascading LDDMM (Wu and Tang, 2018) were employed. Three decreasing $\alpha$ values (0.01, 0.005, 0.002), a multi-resolution scheme and the Cross-Correlation metric were employed in LDDMM. The obtained transformations were then applied to the template labels. The sizes and centers of the patches were calculated using the patch center and size determination method. As depicted in **Figure 2**, a dilation operation was conducted after the union operation. Considering the superior performance of affine+LDDMM, the union region can generally contain the target ROI. To sufficiently limit the patch size, the radius of dilation was set to be 3.

At the training stage, for each ROI, a total of 1000 training patches were extracted. To fully capture the ROI's features, the training patches were extracted from three parts: 1) the patches whose centers are inside the ROI, 2) the patches whose centers are on the boundary of the ROI, 3) the patches that are uniformly distributed at a step size of $\lceil \boldsymbol{R}/10 \rceil$ (where $\boldsymbol{R}$ is the patch size, $\lceil \cdot \rceil$ is the round up operator) in the cuboid containing the ROI. The center sampling approach is depicted on the left of **Figure 3**. As illustrated on the right of **Figure 3**, $K$ most similar atlas patches for each training patch were identified from the transformed template atlases after determining the patch centers. To identify the most similar patches, we set the searching radius to be $0.75\boldsymbol{R}$. In addition, we set $K = 3$ in this work.

In the proposed 3D S-FCN architecture (**Figure 4**), there are a total of 22 layers including 3 layers for multi-encoding and 19 layers for encoding-decoding. The 2 convolutional layers of the multi-encoding part respectively involved 32 and 64 feature maps. Each max pooling layer in the proposed architecture uses $2 \times 2 \times 2$ filters. The number of feature maps of each layer in encoding is respective 256, 64, 128, 128, 256, 256 and 512, and the kernel size of these layers is $3 \times 3 \times 3$ with a stride size of 1. In the decoding part, since an application of long skip connection strategy, the number of feature maps of each layer in decoding is respective 768, 256, 256, 384, 128, 128, 192, 64, 64 and 2. Two small kernels of size $3 \times 3 \times 3$ and $2 \times 2 \times 2$ were used alternatively in these layers. At the end of S-FCN, the probability of a specific ROI was obtained using the softmax function. Optimization of the network parameters was performed using stochastic gradient descent (SGD). Cross-entropy and ReLU (Krizhevsky et al., 2012b) were used as the cost function and the activate function. To further reduce the impact of class imbalance, we used the Dice similarity coefficient (DSC) (Dice, 1945) to determine the ending time of the training procedure.

After determining the above basic elements of S-FCN, an ensemble model, M-FCN, consists of $n$ S-FCN was constructed. In this work, 3 S-FCNs were used to construct the M-FCN. As mentioned in Section 2.4, we adopted a random initialization of the weights and a random shuffling strategy for 3 times for each of these 3 S-FCN models. In this strategy, the weights in layer $l$ were initialized based on a uniform distribution $U[-1/\sqrt{n_l}, 1/\sqrt{n_l}]$, where $n_l$ denotes the number of connections to units in that layer. In our architecture, we set the dropout probability to be 0.1. To prevent the training results from getting stuck into local minima, we firstly performed 5 training epochs to get the maximum DSC and then terminated the training process when this value decreased. For every ROI, the total epochs for training were usually less than 10 with a batch size of 2.

We developed our 3D M-FCN architecture using PyTorch (https://github.com/pytorch). The facility used for training is an Intel(R) Xeon(R) CPU E5-2695 v4 2.10GHz CPU, equipped with two NVIDIA GeForce GTX TITAN XP GPU with 12 GB of memory.

### 2.5.3 Evaluation

To quantify the segmentation accuracy, we used the volume-based metric, DSC, as the evaluation measure. Let $V_{man}$ and $V_{auto}$ denote the manual segmentation and the corresponding automatic segmentation, DSC is defined as

$$DSC(V_{auto}, V_{man}) = \frac{2|V_{auto} \cap V_{man}|}{|V_{auto}| + |V_{man}|}, \tag{10}$$

where $V_{auto} \cap V_{man}$ denotes the intersection of $V_{auto}$ and $V_{man}$, and $|\cdot|$ indicates the volume measurement of the image.

# 3 Experimental Results

## 3.1 Results of comparing different FCN architectures

Two important components of the proposed architecture, as described in Section 2, are the multi-atlas guide learning and long skip connection. To quantitatively assess the impact of these two strategies, we compare the automated segmentation results of S-FCN with that of another three architectures on the first dataset. The other three architectures include: 1) a basic architecture which is part of S-FCN without the $K$ similar atlas patches convolutional layers nor skip connections, named S-FCN (base); 2) the multi-atlas guided S-FCN that adds the $K$ similar atlas patches convolutional layers to the beginning of S-FCN (base), named S-FCN (multi-atlas); 3) the architecture that uses the skip connection strategy in S-FCN (base), named S-FCN (skip). To conduct a fair comparison, all other configurations of these four networks are the same as that described in Section 2.3.

**Table 1** presents the mean and standard derivations of the DSC for each of the 14 ROIs as well as the overall DSC (all labels) obtained from the four networks. Evidently, each DSC value of the proposed S-FCN architecture is no lower than that of the other three ones. According to paired Student's $t$-tests, the DSC values obtained from the proposed S-FCN are statistically higher than those from the other three architectures for all 14 ROIs except the right pallidus, the left thalamus and the left amygdala ($p < 0.01$). Overall, the proposed S-FCN works the best. In addition, the standard derivatives of S-FCN are lower than that of the other three architectures (0.013 for S-FCN, 0.019, 0.018 and 0.021 for S-FCN (base), S-FCN (multi-atlas) and S-FCN (skip)).

Employing either multi-atlas guide learning (S-FCN (multi-atlas)) or long skip connection (S-FCN (skip)) benefits the segmentation accuracy; the DSC values obtained from either S-FCN (multi-atlas) or S-FCN (skip) are higher than those from S-FCN (base) for a majority of the 14 ROIs. Compared to S-FCN (base), the improvement of the long skip connection strategy is larger than that of the multi-atlas guided learning strategy (0.023 versus 0.003 overall). Despite the slight improvement of the multi-atlas guided strategy, when combining the two strategies, the segmentation performance is further boosted (0.063). It clearly demonstrates the importance of combining those two strategies in the proposed S-FCN architecture.

## 3.2 Results of comparing different segmentation methods

To validate the performance of the proposed method (M-FCN), we compared its segmentation performance with that of another three segmentation methods: joint Label fusion (JLF) (Wang and Yushkevich, 2013), LiviaNET (Dolz et al., 2018) and the proposed S-FCN. JLF represents the state-of-the-art weighted label fusion method. To ensure the fairness of comparison, the registration step in JLF is the same as that described in Section 2.5.2. Also, the parameters used are the same as those suggested in Wang and Yushkevich (2013) with $\alpha = 0.1$, $\beta = 2$, $r_p = 2$, and $r_s = 3$.

LiviaNET is a representative 3D FCN model, which has been suggested to perform well in subcortical structure segmentation. LiviaNET consists of 13 layers with each convolutional layer using the same number of kernels of size $3 \times 3 \times 3$. It was trained for 30 epochs each of which was again composed of 20 sub-epochs. At each sub-epoch, a total of 500 samples were randomly selected from the training image segments, and processed in the batch size of 5. Since LiviaNET was proposed to segment the subcortical structures and our second dataset has other brain regions defined, the comparisons involving LiviaNET were only conducted on the first dataset.

### 3.2.1 Experimental results on the first dataset

For the first dataset, the mean and standard deviations of the DSC values, for all 14 ROIs, produced by JLF, LiviaNET, S-FCN, and M-FCN are presented in **Table 2**. The mean DSC values of our proposed methods including both S-FCN and M-FCN are higher than those of the other two methods for all of the 14 ROIs. According to paired Student's $t$-tests , M-FCN is statistically significantly better than the other three methods in segmenting 8 ROIs including the right caudate, the left putamen, the binary thalamus, the left amygdala, the left hippocampus and the bilateral lateral ventricle ($p < 0.05$). Comparing M-FCN and JLF, M-FCN performs significantly better than JLF in segmenting 13 ROIs (except the right amygdala), as evaluated by the DSC value ($p < 0.05$), especially for the left thalamus, the left hippocampus and the left lateral ventricle ($p < 2.4e^{-5}$). Comparing M-FCN and LiviaNET, M-FCN performs statistically significantly better for all 14 ROIs ($p < 0.05$), especially for the left hippocampus, the bilateral thalamus and the bilateral lateral ventricle ($p < 4.7e^{-4}$).

**Figure 6** shows the distributions of the DSC values of the 16 subjects, obtained from the 4 segmentation methods, for each of the 14 ROIs. It is clear that both S-FCN and M-FCN are superior to the other two methods. The DSC ranges of JLF, LiviaNET, S-FCN and M-FCN for all 14 ROIs are respectively $[0.689, 0.952]$, $[0.438, 0.950]$, $[0.743, 0.963]$ and $[0.758, 0.964]$. A smaller interval of M-FCN's DSC range, but higher mean DSC values, compared to S-FCN demonstrates the ensemble strategy's capability of improving the segmentation robustness. A visual comparison of the segmentation results between the proposed method and the other 3 methods of one representative subject from the first dataset are shown in **Figure 7**. Unlike those of JLF and LiviaNET, the segmentation results of the proposed S-FCN and M-FCN do not contain undesired isolated regions, which indicates the superiority of the proposed architecture in maintaining spatial consistency. Compared with the results of S-FCN, the boundaries generated by M-FCN are more smooth, indicating that the ensemble strategy and the averaging operation are beneficial for classifying voxels on the boundary.

### 3.2.2 Experimental results on the second dataset

For the second dataset, the mean and standard derivations of the DSC values for all 54 ROIs as well as the overall DSC obtained from each of the 3 segmentation methods are listed in **Table 3**. The proposed M-FCN performs the best in segmenting all ROIs except for the right supramarginal gyrus, the left angular gyrus and the left Cingulate gyrus. For a total of 28 ROIs, M-FCN has significantly higher DSC values than both JLF and S-FCN ($p < 0.05,$ paired $t$-test).

As demonstrated in **Table 3**, for a total of 6 subcortical structures (left and right caudate, left and right putamen as well as left and right hippocampus), M-FCN is significantly superior to JLF, which is the same for the first dataset. For the other ROIs of larger volume size, such as the frontal gyrus, the segmentation performance of M-FCN is also better than that of JLF. It suggests that the proposed M-FCN delivers competitive segmentation results not only for subcortical structures but also for other brain ROIs. Distributions of the DSC values of the second dataset, obtained from each of the 3 methods for comparison, are depicted in **Figure 8**. The distribution of the DSC values from M-FCN is more concentrated than that of S-FCN, indicating again the effectiveness of the ensemble strategy.

### 3.2.3 Computation time

It generally takes about 20 minutes to fully train the network of M-FCN for a single ROI with the patch size being $25 \times 60 \times 40$. The total training and testing time may be different depending on the number of atlases, the number of ROIs and the size of the images. For the first dataset, the training time is around 17 hours, and the testing time per image is around 35 seconds. For the second dataset, the training time is around 51 hours, and the testing time per image is around 2 minutes. All methods for comparison are evaluated using the same platform. JLF needs around 15 minutes per testing image, and LiviaNET needs around 60 hours for training and around 9 minutes per testing image. Please note the running time for preprocessing is not considered in this section.

# 4 Discussion

In this paper, we have proposed a fully automated method for segmenting brain structures from T1-weighted images based on a multi-atlas guided 3D FCN ensemble model. There are four steps involved in the proposed method: (1) histogram matching and affine+LDDMM registration, (2) determining the size and center of each ROI patch, (3) extracting ROI patch and identifying the corresponding $K$ most similar atlas patches, and (4) segmenting and fusing ROI patches to get the final segmentation map. The effectiveness of multi-atlas guided learning and long skip connection in the proposed FCN architecture for segmenting MRIs has been systematically validated. Furthermore, we investigated the segmentation performance of an ensemble technique based S-FCN (M-FCN), the one exhibiting a superior performance to S-FCN, by comparing it with another two state-of-the-art fully-automatic segmentation methods, JFL and LiviaNET, in terms of both segmentation accuracy and computational efficiency. To validate the robustness of our findings, all experiments were conducted on two different datasets of brain images involving different numbers of ROIs.

Compared to other state-of-the-art segmentation techniques relying on patch and multi-atlas method, JLF generally has an outstanding segmentation accuracy. As such, in this paper, we compared the experimental results of the proposed method with that of JLF. On both datasets, our method exhibited a superior segmentation performance over JLF, especially for subcortical structures. The reason for the relatively lower accuracy of JLF is that a coarse image registration may induce unsatisfactory segmentation results. Furthermore, similar to other patch-based methods, because of a relatively small patch size, only local features of the ROIs are learned in the fusion process which may have also affected the segmentation accuracy. On the contrary, our proposed method uses a dilated patch surrounding each single ROI to conduct segmentation, and thus both local and global features, especially the boundary information of each ROI can be efficiently learned.

For CNN based segmentation methods, their architectures can generally be divided into two types; basing on either 2D or 3D networks. Compared to 3D approaches, an obvious advantage of a 2D approach is its lower computational and memory requirement. However, the main drawback of such 2D methods is their complete discarding of anatomic information along the direction orthogonal to the 2D plane. Since brain MRIs are typically 3D, apparently, the context expressed in the form of 3D is more abundant than in 2D. Numerous strategies for improving the performance of 2D architectures have been proposed (de Brebisson and Montana, 2015; Shakeri et al., 2016). Another strategy to fully excavate the context of 3D volumes is to employ 3D architectures. LiviaNET is a representative 3D FCN architecture for subcortical structure segmentation. In this method, registration is not needed, and based on a large-scale study, its ability to capture a wide range of variability has been identified. Although LiviaNET can learn the variability context of different structures across large-scale unregistered MRIs, it is prone to isolated labels (as shown in **Figure 7**), which may have been due to the limited dataset size.

Another potential reason of the isolated labels is that a CNN architecture utilizing only intensity and patches of fixed size may lack the ability to discriminate voxels with similar intensity profile within different ROIs. According to our experimental results, employing patches of adaptive sizes for different ROIs can significantly improve the network performance when segmenting a challenging boundary region. Our patch extraction method can introduce prior location information of different ROIs into the CNN network. As such, only tightly-related local and global features of each ROI were learned in the segmentation process, which largely reduced the training time for learning background samples, and therefore reduced the false negative (see **Figure 7**).

Another main contribution of this work is the proposed multi-atlas guided method. Being different from intensity based FCN methods (de Brebisson and Montana, 2015; Shakeri et al., 2016; Dolz et al., 2018), it also introduces the label information of most similar patches into the network learning process. And thus, the anatomical shape and structure patterns are taken into account when assigning a label to a voxel. Being different from the multi-atlas guided framework presented in Fang et al. (2017), a long skip connection strategy, ensemble technique and LDDMM registration are employed in our proposed method. The long skip connection strengthens the ability of S-FCN for learning the global context, the ensemble technique improves the performance for accurately identifying different ROIs' boundaries, and the LDDMM registration ensures that the patches extracted contain the to-be-segmented ROI even with a small size (a small dilation radius).Typically limited by the dataset size, simultaneously segmenting a large number of ROIs is still a challenging problem; the error rate will generally increase when the number of ROIs for segmentation increases, especially for intensity-only based FCN architectures. Thus, previous CNN based segmentation studies generally deal with subcortical structures (Dolz et al., 2018), or some other simple structures (Zhang, et al., 2015). On the contrary, in this paper, our proposed method shows an outstanding performance even for segmenting a large number of ROIs (whole brain). And such high segmentation accuracy does not require extra training samples, which further demonstrates the importance and effectiveness of our multi-atlas guided scheme and the aforementioned three strategies.

One potential limitation of this work is that we did not consider the memory footprint in training the network for segmenting ROIs of large size. Combining the proposed framework with the dense training strategy presented elsewhere (Dolz et al., 2018) for large ROIs and using patches of small size may reduce the memory footprint of the proposed method without scarifying the segmentation accuracy. This will be one of our future endeavors.

# 5 Conclusions

In this paper, a new multi-atlas guided 3D M-FCN method has been proposed for brain ROI labeling. Compared to traditional neural networks designed for brain segmentation, the proposed 3D FCN network learns not only the image information but also the label information. Moreover, by employing patches of adaptive sizes in training, we can obtain a deep network for fast training and testing. More importantly patch based segmentation can effectively reduce isolated labels. The ROI-specific local and global contexts are modeled by long skip connection between the layers of encoding and decoding parts, encouraging a superposition of consistent features. Utilizing ensemble techniques further reduce overfitting and enhance the segmentation robustness. The proposed approach demonstrates superior performance to several state-of-the-art segmentation algorithms on two MRI datasets involving different brain regions. Significant improvements have been observed in terms of the overall segmentation accuracy as well as the segmentation accuracy of each single ROI.

## Acknowledgment

This study was supported by the National Natural Science Foundation of China (NSFC 81501546) and the National Key R&D Program of China (2017YFC0112404).

**Table 1**

Mean and standard derivation of the DSC values obtained from the four tested S-FCN architectures for all 14 ROIs of the first dataset. Bold typesetting indicates that the DSC value obtained from the corresponding method is statistically significantly higher than those of the other three methods ($p<0.01$).

| | S-FCN (base) | S-FCN (multi-atlas) | S-FCN (skip) | S-FCN |
|---|---|---|---|---|
| Right Caudate | 0.858 (0.016) | 0.878 (0.016) | 0.861 (0.025) | **0.936 (0.016)** |
| Left Caudate | 0.884 (0.017) | 0.876 (0.013) | 0.803 (0.046) | **0.930 (0.018)** |
| Right Pallidum | 0.865 (0.028) | 0.871 (0.028) | 0.885 (0.005) | 0.887 (0.024) |
| Left Pallidum | 0.871 (0.017) | 0.855 (0.044) | 0.879 (0.013) | **0.888 (0.021)** |
| Right Putamen | 0.847 (0.029) | 0.892 (0.026) | 0.928 (0.010) | **0.934 (0.010)** |
| Left Putamen | 0.858 (0.027) | 0.855 (0.021) | 0.912 (0.025) | **0.934 (0.013)** |
| Right Thalamus | 0.916 (0.078) | 0.915 (0.071) | 0.930 (0.032) | **0.943 (0.022)** |
| Left Thalamus | 0.924 (0.067) | 0.935 (0.031) | 0.939 (0.021) | 0.940 (0.017) |
| Right Amygdala | 0.795 (0.037) | 0.837 (0.027) | 0.807 (0.034) | **0.875 (0.023)** |
| Left Amygdala | 0.800 (0.013) | 0.808 (0.018) | 0.873 (0.022) | 0.873 (0.016) |
| Right Hippocampus | 0.855 (0.019) | 0.848 (0.026) | 0.906 (0.018) | **0.915 (0.013)** |
| Left Hippocampus | 0.837 (0.020) | 0.847 (0.021) | 0.830 (0.042) | **0.916 (0.014)** |
| Right Lateral Ventricle | 0.708 (0.108) | 0.754 (0.039) | 0.763 (0.105) | **0.866 (0.052)** |
| Left Lateral Ventricle | 0.736 (0.128) | 0.770 (0.047) | 0.832 (0.087) | **0.881 (0.051)** |
| All labels | 0.850 (0.019) | 0.853 (0.018) | 0.876 (0.021) | **0.917 (0.013)** |

**Table 2**

Mean and standard derivation of the DSC values for each of the four methods of comparison on every ROI of the first dataset. Bold typesetting indicates that the DSC value obtained from the corresponding method is statistically significantly higher than those of the other three methods ($p<0.05$).

| | JLF | LiviaNET | S-FCN | M-FCN |
|---|---|---|---|---|
| Right Caudate | 0.922 (0.026) | 0.922 (0.043) | 0.936 (0.016) | **0.942 (0.012)** |
| Left Caudate | 0.918 (0.020) | 0.922 (0.012) | 0.930 (0.018) | 0.935 (0.013) |
| Right Pallidum | 0.886 (0.025) | 0.865 (0.030) | 0.887 (0.024) | 0.893 (0.024) |
| Left Pallidum | 0.887 (0.016) | 0.872 (0.030) | 0.888 (0.021) | **0.899 (0.018)** |
| Right Putamen | 0.933 (0.008) | 0.917 (0.037) | 0.934 (0.010) | 0.937 (0.009) |
| Left Putamen | 0.928 (0.009) | 0.915 (0.048) | 0.934 (0.013) | 0.936 (0.014) |
| Right Thalamus | 0.924 (0.018) | 0.898 (0.030) | 0.943 (0.022) | **0.945 (0.022)** |
| Left Thalamus | 0.933 (0.014) | 0.916 (0.022) | 0.940 (0.017) | **0.943 (0.015)** |
| Right Amygdala | 0.867 (0.020) | 0.812 (0.119) | 0.875 (0.023) | 0.877 (0.023) |
| Left Amygdala | 0.861 (0.023) | 0.831 (0.051) | 0.873 (0.016) | **0.878 (0.017)** |
| Right Hippocampus | 0.905 (0.013) | 0.884 (0.037) | 0.915 (0.013) | 0.917 (0.012) |
| Left Hippocampus | 0.905 (0.013) | 0.865 (0.047) | 0.916 (0.014) | **0.919 (0.012)** |
| Right Lateral Ventricle | 0.839 (0.069) | 0.837 (0.051) | 0.866 (0.052) | **0.871 (0.049)** |
| Left Lateral Ventricle | 0.843 (0.070) | 0.839 (0.059) | 0.880 (0.051) | **0.888 (0.048)** |
| All labels | 0.904 (0.010) | 0.889 (0.018) | 0.917 (0.013) | **0.922 (0.012)** |

**Table 3**

Mean and standard deviations of the DSC values for each of the 54 whole brain ROIs of the second dataset, obtained from JLF, S-FCN, and M-FCN. Bold typesetting indicates that the DSC value obtained from the corresponding method is statistically significantly higher than those of the other two methods (p<0.05).

| | **JLF** | | **S-FCN** | | **M-FCN** | |
|---|---|---|---|---|---|---|
| | Left | Right | Left | Right | Left | Right |
| sup. frontal gyrus | 88.2 (2.0) | 87.9 (2.1) | 88.8 (2.3) | 88.4 (2.4) | **89.1 (2.3)** | **89.3 (2.2)** |
| middle frontal gyrus | 86.4 (3.2) | 86.1 (3.1) | 87.3 (3.2) | 85.6 (3.6) | **87.8 (3.3)** | 86.7 (3.5) |
| inf. frontal gyrus | 82.2 (4.3) | 81.2 (5.4) | 82.8 (4.8) | 80.2 (6.4) | **83.5 (4.6)** | 81.6 (6.5) |
| precentral gyrus | 86.6 (2.8) | 85.6 (3.0) | 87.4 (5.0) | 86.6 (3.5) | **88.3 (3.2)** | **87.2 (3.4)** |
| middle orbitofrontal gyrus | 77.5 (6.4) | 77.5 (6.1) | 76.4 (8.1) | 76.9 (6.7) | 77.6 (7.7) | 77.5 (6.8) |
| lateral orbitofrontal gyrus | 71.2 (5.9) | 70.2 (8.3) | 72.5 (8.3) | 70.8 (9.3) | **74.6 (7.3)** | 71.5 (8.8) |
| gyrus rectus | 78.9 (4.9) | 80.4 (5.2) | 80.4 (5.0) | 82.2 (5.2) | **81.2 (4.8)** | **83.0 (4.6)** |
| postcentral gyrus | 82.8 (4.2) | 82.6 (3.8) | 84.0 (7.1) | 82.8 (5.5) | **84.5 (7.1)** | **84.0 (5.0)** |
| sup. parietal gyrus | 83.5 (2.9) | 82.7 (3.5) | 83.6 (3.8) | 82.5 (4.0) | **84.8 (3.4)** | 82.8 (4.0) |
| supramarginal gyrus | 78.7 (5.6) | 78.1 (6.7) | 79.3 (5.9) | 76.7 (7.4) | 80.0 (5.7) | 78.0 (6.9) |
| angular gyrus | 75.9 (6.0) | 76.9 (4.8) | 73.4 (7.5) | 78.1 (4.6) | 74.9 (7.1) | 78.2 (4.3) |
| precuneus | 77.9 (3.5) | 78.7 (3.2) | 78.3 (4.9) | 79.7 (4.5) | 79.1 (4.9) | 79.6 (4.6) |
| sup. occipital gyrus | 72.8 (6.6) | 70.4 (8.1) | 74.8 (6.5) | 71.9 (7.4) | **75.8 (6.4)** | 73.2 (5.4) |
| middle occipital gyrus | 78.4 (4.7) | 77.8 (5.3) | 78.2 (5.3) | 77.6 (5.5) | 79.3 (5.1) | 78.6 (5.5) |
| inf. occipital gyrus | 78.5 (4.7) | 79.2 (3.9) | 78.3 (5.4) | 78.8 (5.7) | **80.3 (5.7)** | 79.7 (5.8) |
| cuneus | 77.6 (6.3) | 77.0 (6.5) | 81.5 (5.2) | 82.1 (4.6) | 81.1 (5.5) | **83.2 (4.4)** |
| sup. temporal gyrus | 85.9 (4.2) | 86.3 (3.0) | 86.6 (4.3) | 87.4 (2.5) | **87.2 (4.2)** | 87.4 (2.6) |
| middle temporal gyrus | 78.9 (4.0) | 80.2 (3.4) | 79.5 (5.1) | 81.5 (3.2) | **80.3 (4.7)** | 81.6 (3.5) |
| inf. temporal gyrus | 78.8 (4.0) | 79.9 (4.6) | 80.0 (5.3) | 80.3 (6.1) | **81.2 (4.9)** | **81.2 (5.3)** |
| parahippocampal gyrus | 82.4 (3.2) | 81.3 (3.8) | 83.9 (3.3) | 82.7 (3.9) | 84.0 (3.2) | **83.5 (3.8)** |
| lingual gyrus | 83.5 (4.0) | 83.9 (4.0) | 84.4 (5.7) | 85.2 (4.6) | **85.0 (5.2)** | **85.7 (4.2)** |
| fusiform gyrus | 83.3 (4.1) | 82.8 (5.2) | 83.8 (4.3) | 82.6 (5.8) | **84.9 (4.1)** | 83.7 (5.8) |
| insular cortex | 87.3 (1.9) | 86.6 (2.3) | 88.5 (2.0) | 88.2 (2.3) | **88.9 (2.1)** | **88.5 (2.1)** |
| cingulate gyrus | 81.0 (3.9) | 79.7 (6.1) | 80.1 (4.4) | 79.5 (5.3) | 79.9 (4.6) | 80.4 (4.5) |
| caudate | 85.5 (3.9) | 84.9 (4.6) | 86.2 (4.9) | 85.1 (5.9) | **86.6 (4.6)** | 85.7 (5.7) |
| putamen | 85.9 (2.3) | 86.1 (2.9) | 86.1 (2.5) | 85.9 (3.1) | 86.3 (2.5) | 86.3 (3.2) |
| hippocampus | 83.9 (3.0) | 83.9 (2.2) | 84.1 (3.0) | 84.3 (2.2) | **84.8 (3.2)** | **84.8 (2.1)** |
| All Labels | 82.6 (1.0) | | 83.1 (1.0) | | **83.8 (1.0)** | |

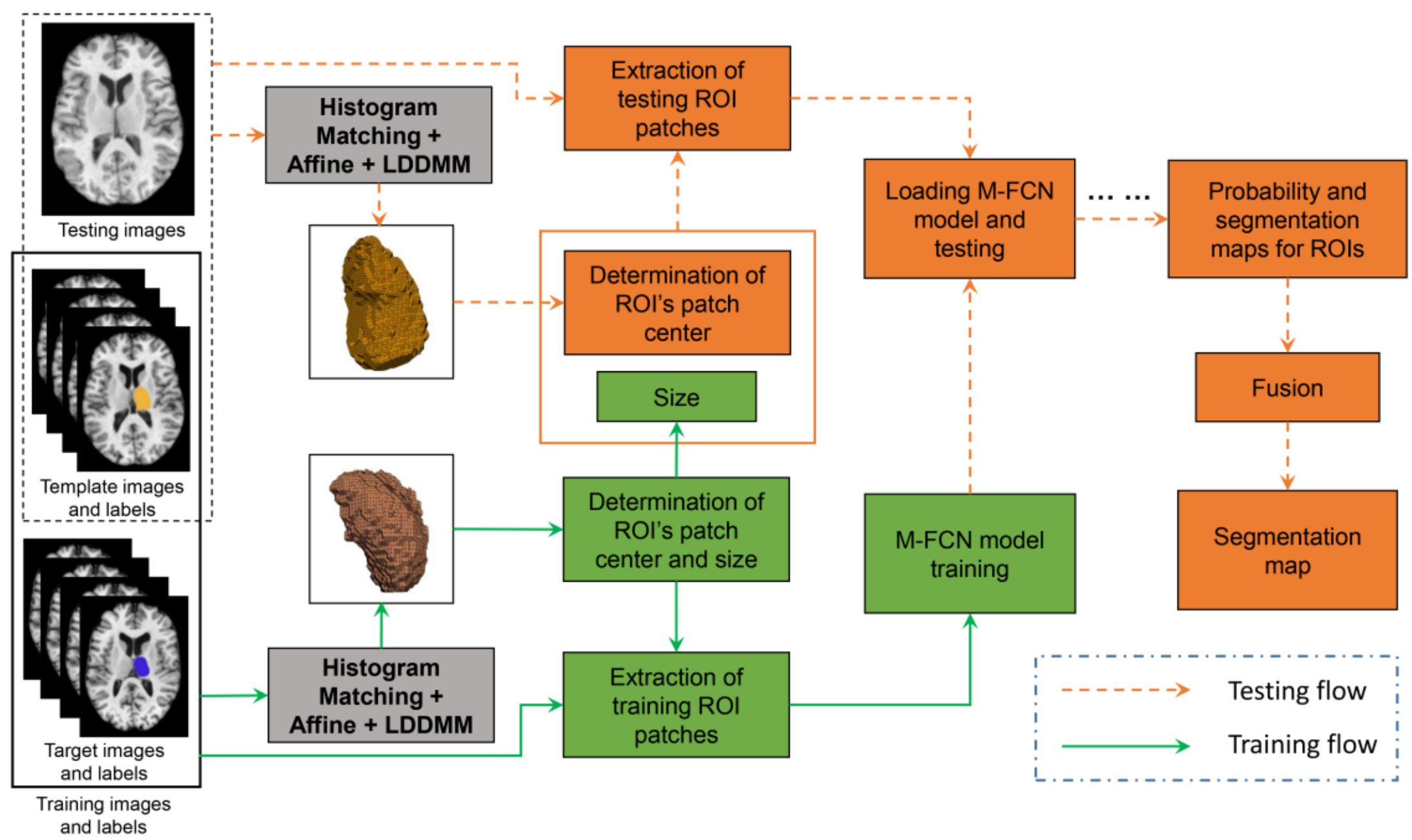

**Figure 1:** Overview of the proposed segmentation framework. MRI dataset are preprocessed for histogram matching and then the template images are aligned to the testing and target images using affine followed by LDDMM registration. The ROI's patch center and size are determined from the transformed template labels. The ROI patches for training and testing are extracted and then the M-FCN model is trained. The extracted testing ROI patch is loaded into the trained M-FCN model to get the label and probability patch. After all ROIs' patches of the testing image are segmented by the M-FCN models, the resultant probability and segmentation maps are then fused to yield the final segmentation map of the testing image.

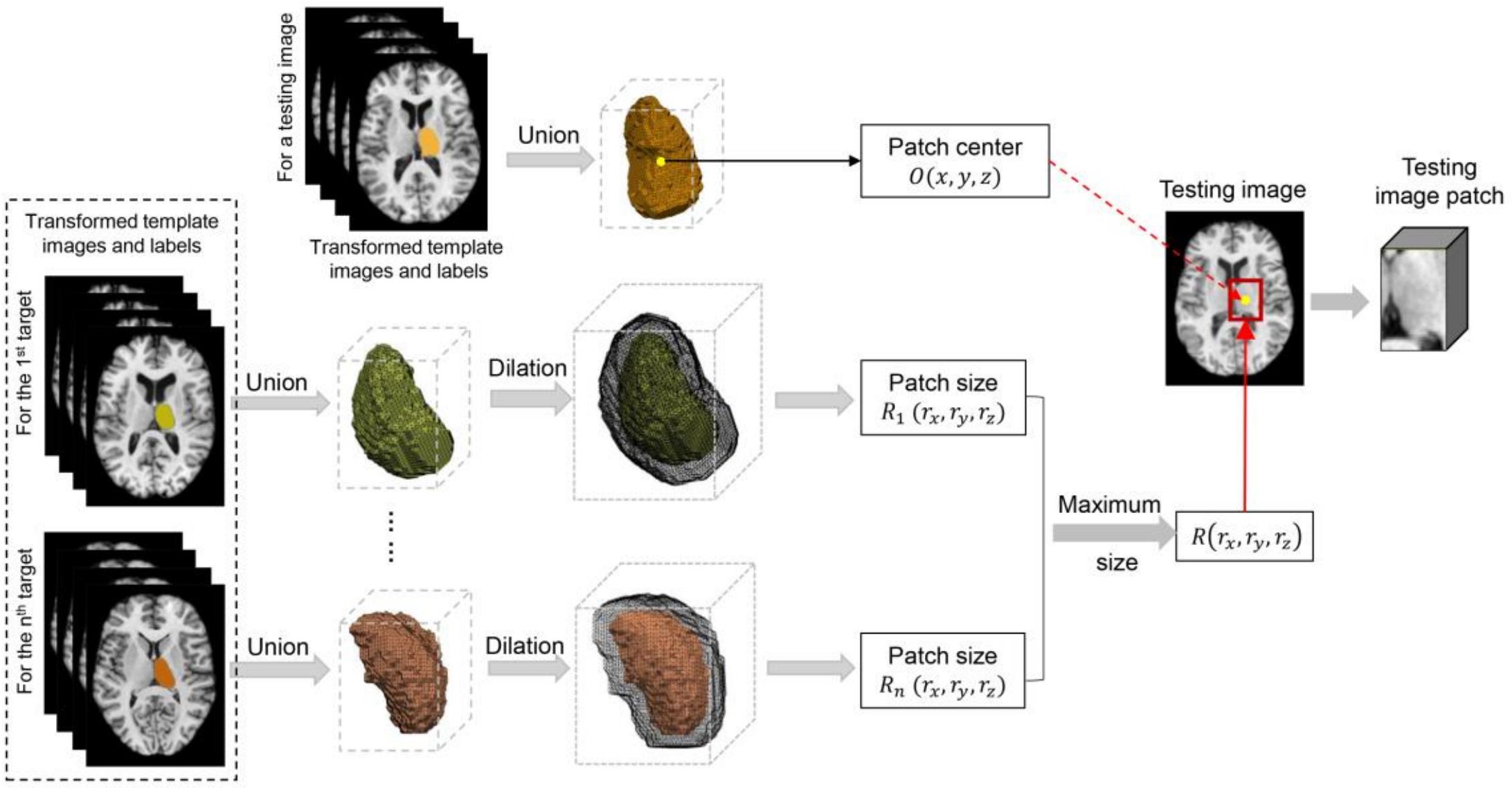

**Figure 2:** Determination of the patch size and center of a ROI for a testing image.

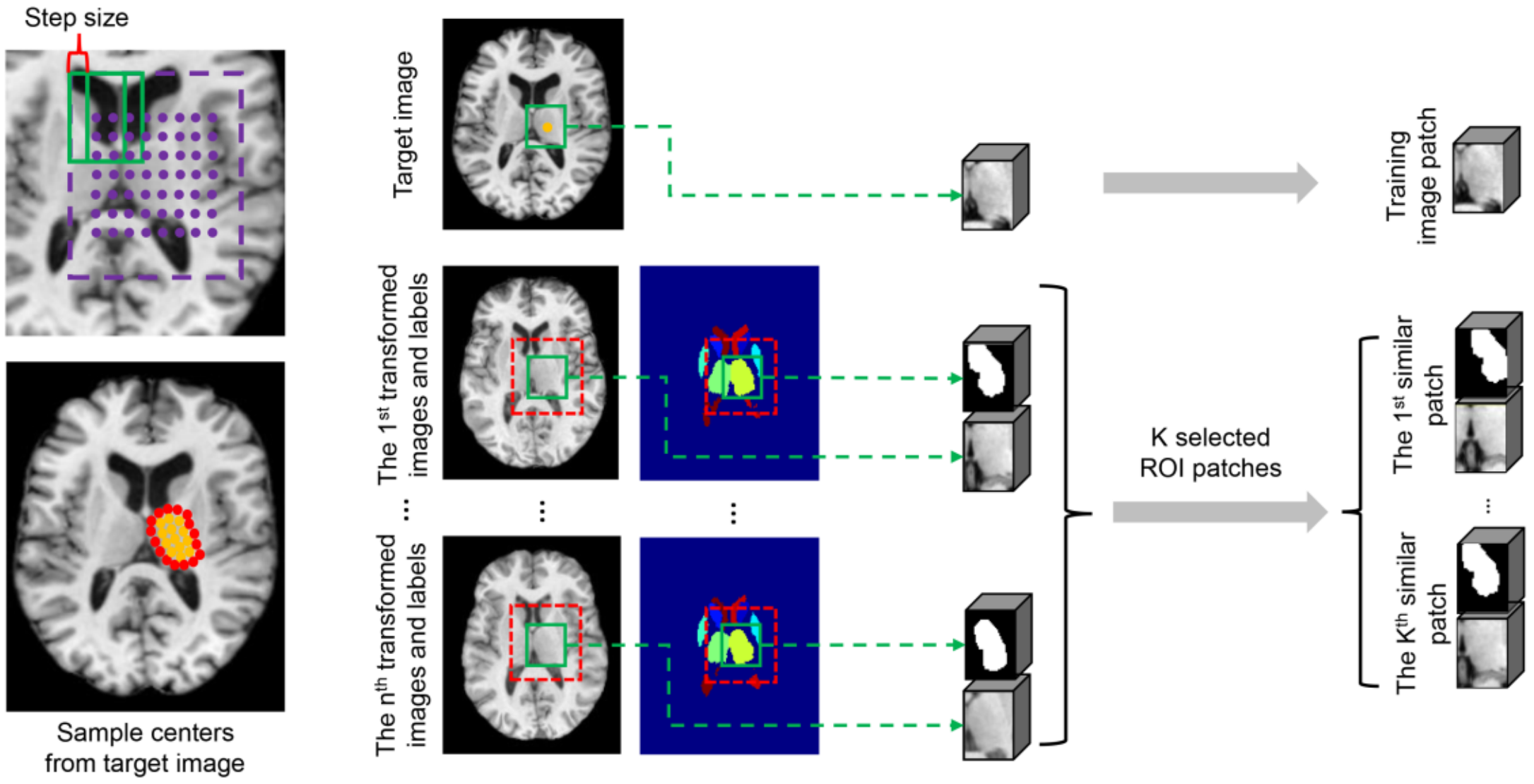

**Figure 3:** Method for extracting image and label patches for training. The ROI's patch centers are firstly sampled from the labels of the target images. The sampled centers consist of three parts: 1) the centers which are uniformly distributed in the region (purple dotted frame) covering the ROI (purple dots), 2) centers inside the ROI (yellow dots), 3) centers on the boundary of the ROI (red dots). The ROI patches centered at the sampled centers are extracted and then *K* most similar ROI image and label patches are identified in a small neighborhood with size *N* (red dotted frame) from the transformed template images and labels. The extracted patches as well as the corresponding *K* most similar patches will be used as the training samples.

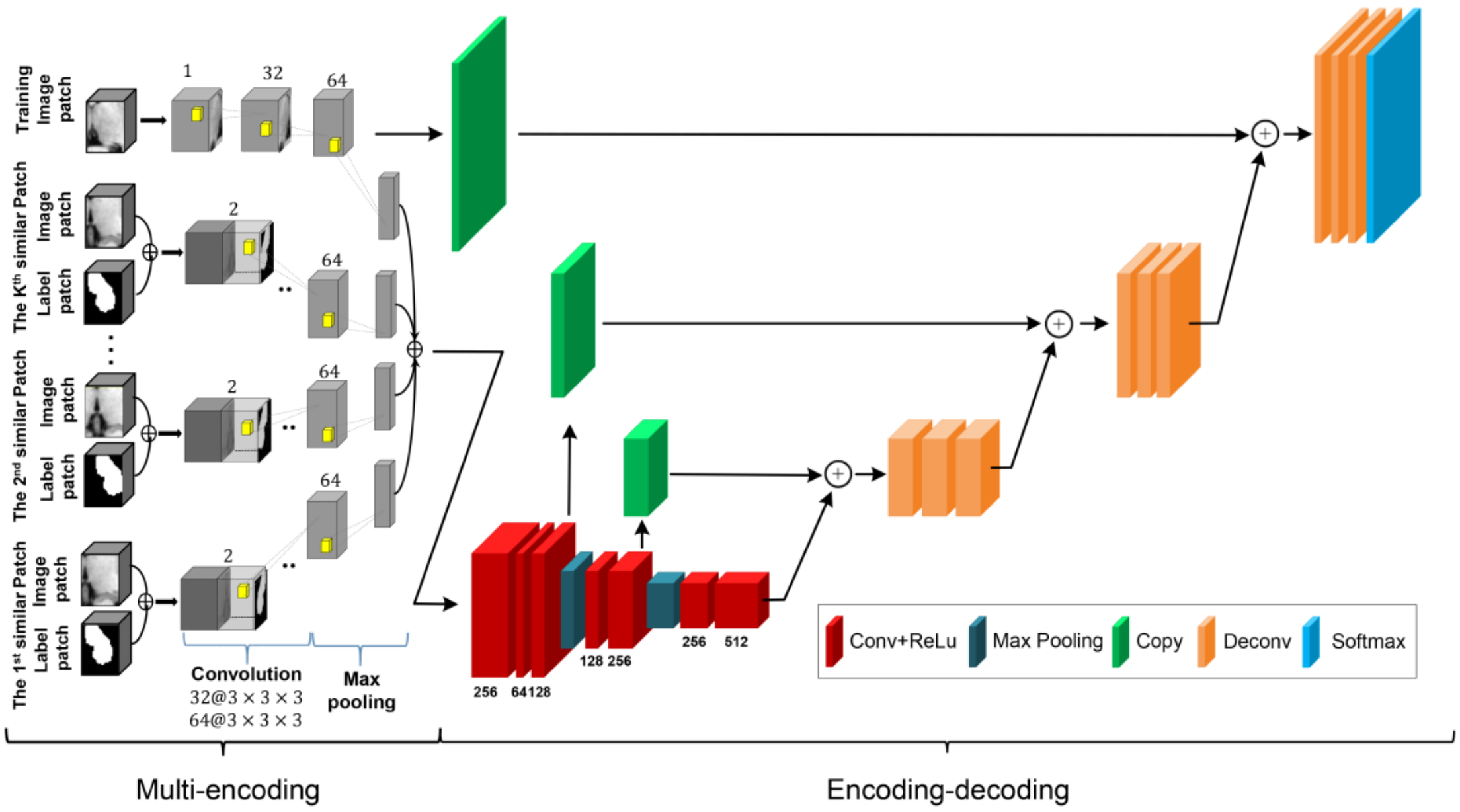

**Figure 4:** The overall framework of the training stage in a single FCN model of the proposed multi-atlas based M-FCN. The training ROI image patch with $K$ most similar patches are fed into the single FCN model for training.

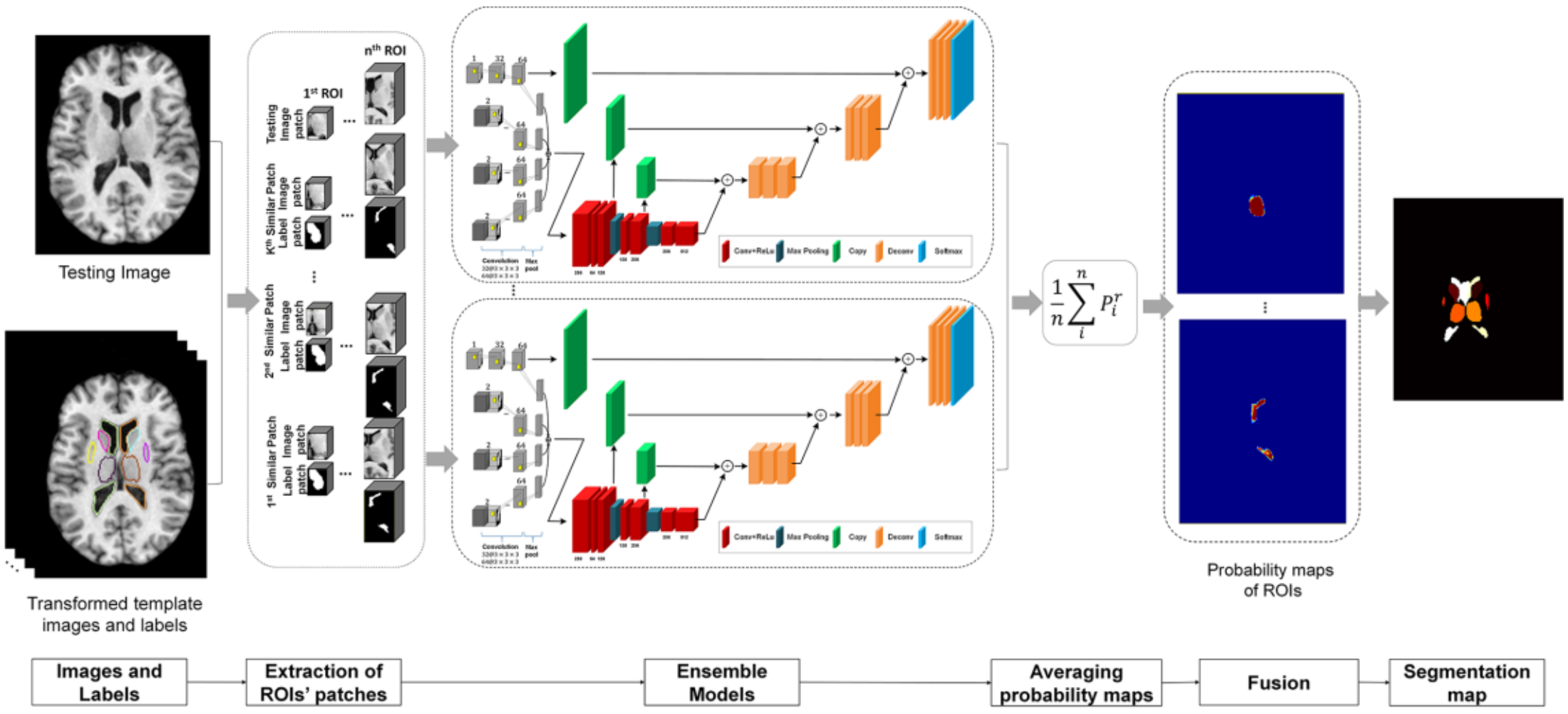

**Figure 5:** The overall framework of the testing stage in the proposed multi-atlas based M-FCN.

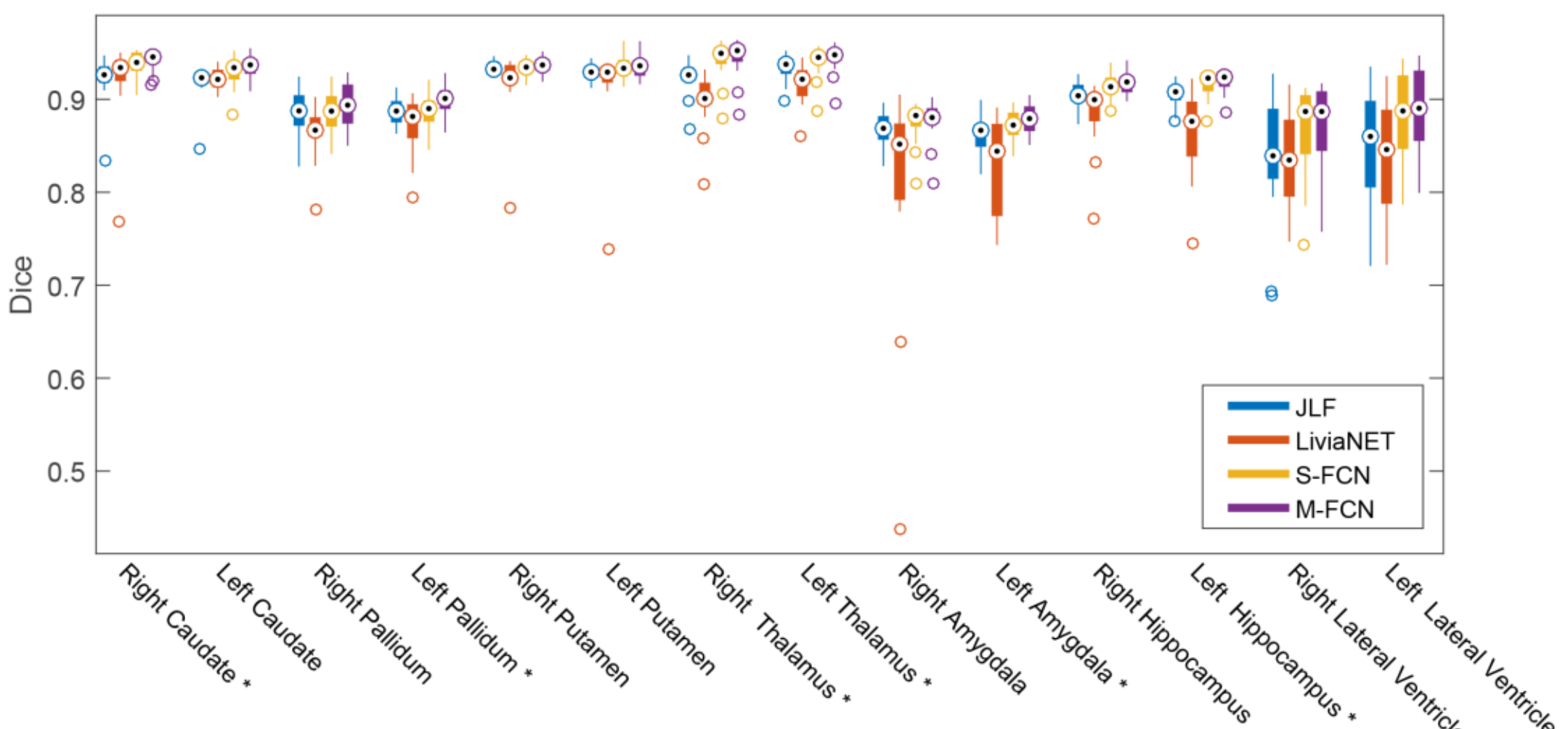

**Figure 6:** Distributions of the DSC values for each of the 14 ROIs of the first dataset obtained from each of the 4 methods. An asterisk indicates that M-FCN is superior to each of the other three methods in segmenting the corresponding ROI.

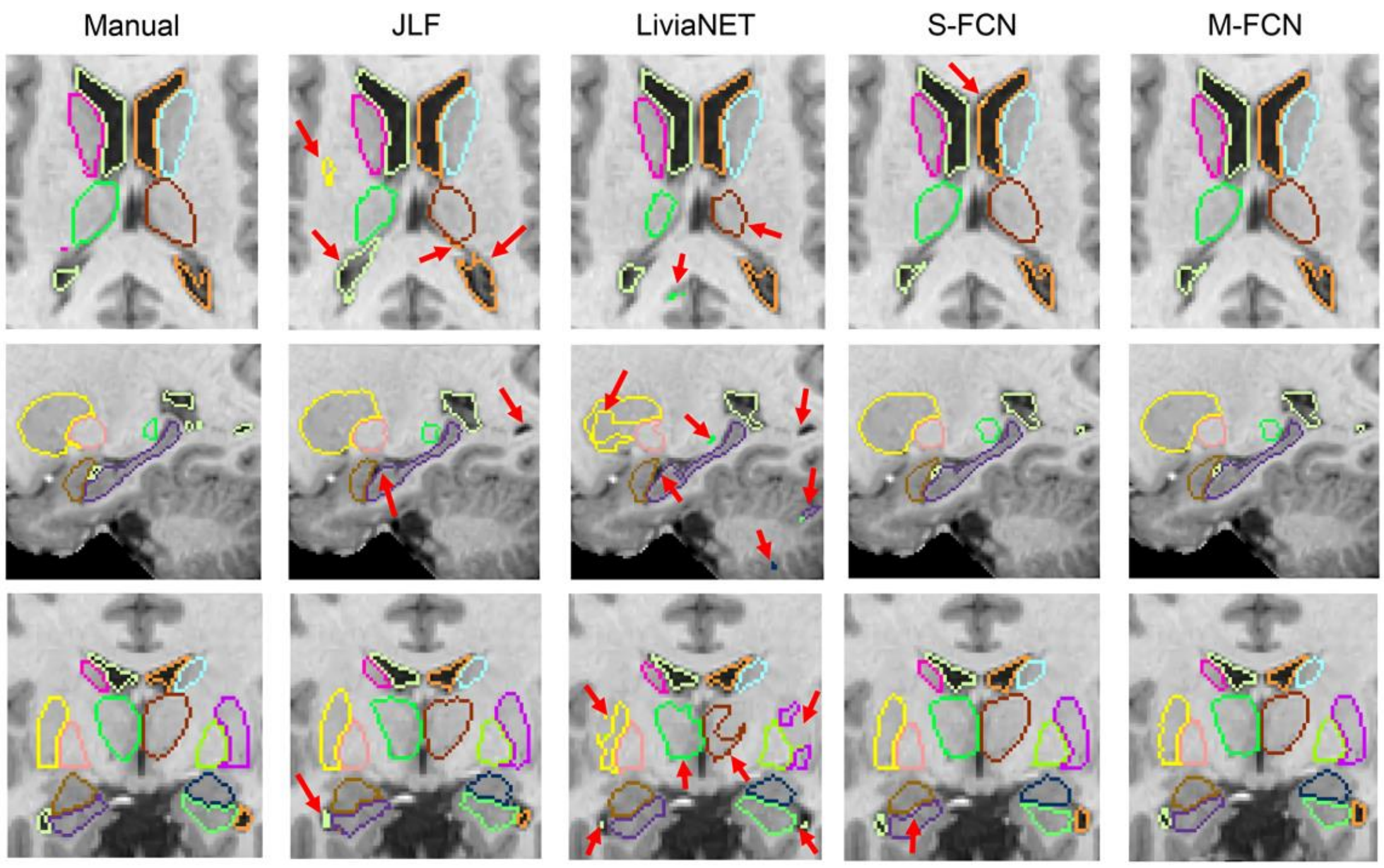

**Figure 7:** A visual comparison of the segmentation results of the 14 ROIs, obtained from the five segmentation methods (manual, JLF, LiviaNET, S-FCN, M-FCN), in one representative subject of the first dataset.

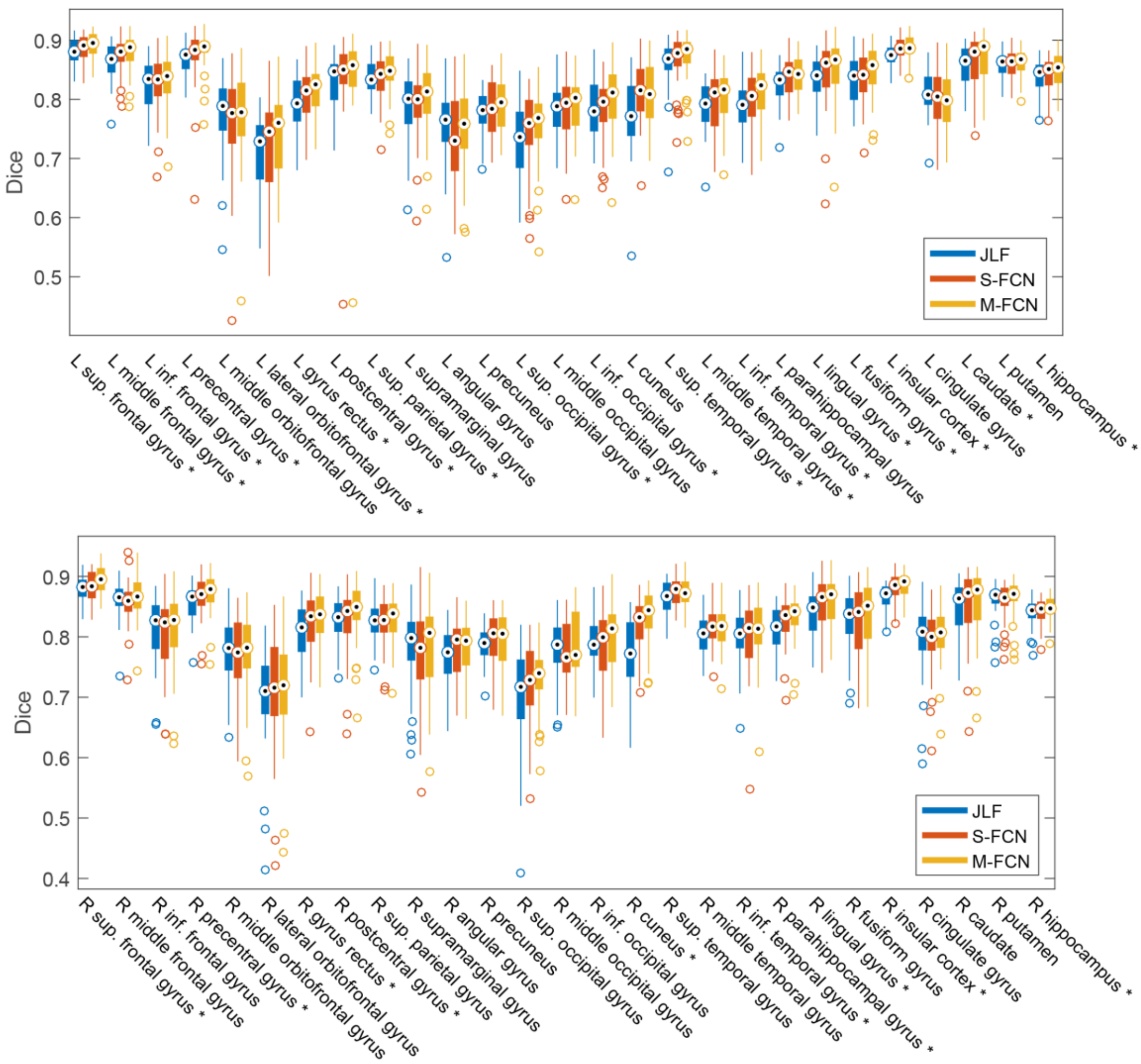

**Figure 8:** Distributions of the DSC values for each of the 54 ROIs of the second dataset obtained from each of the 3 methods. An asterisk indicates that M-FCN is superior to each of the other two methods in segmenting the corresponding ROI.